\documentclass[10pt,twocolumn,letterpaper]{article}

\usepackage{cvpr}
\usepackage{times}
\usepackage{epsfig}
\usepackage{graphicx}
\usepackage{amsmath}
\usepackage{amssymb}
\usepackage{booktabs}
\usepackage{float}
\usepackage{xcolor}
\usepackage{tabulary}
\usepackage{multirow}
\usepackage[caption=false]{subfig}

\usepackage[british,english,american]{babel}

\newcommand{\tablestyle}[2]{\setlength{\tabcolsep}{#1}\renewcommand{\arraystretch}{#2}\centering\footnotesize}
\newlength\savewidth\newcommand\shline{\noalign{\global\savewidth\arrayrulewidth
  \global\arrayrulewidth 1pt}\hline\noalign{\global\arrayrulewidth\savewidth}}
\newcolumntype{x}[1]{>{\centering\arraybackslash}p{#1pt}}
\newcommand{\bftab}{\fontseries{b}\selectfont}
\newcommand{\bd}[1]{\textbf{#1}}

\newcommand{\app}{\raise.17ex\hbox{$\scriptstyle\sim$}}

\definecolor{demphcolor}{RGB}{144,144,144}
\newcommand{\demph}[1]{\textcolor{demphcolor}{#1}}

\definecolor{citecolor}{RGB}{34,139,34}
\usepackage[pagebackref=true,breaklinks=true,letterpaper=true,colorlinks,
  citecolor=citecolor,bookmarks=false]{hyperref}

\makeatletter\renewcommand\paragraph{\@startsection{paragraph}{4}{\z@}
  {.5em \@plus1ex \@minus.2ex}{-.5em}{\normalfont\normalsize\bfseries}}\makeatother

 \cvprfinalcopy 

\def\cvprPaperID{****}

\ifcvprfinal\fi
\begin{document}

\title{Data Distillation: Towards Omni-Supervised Learning \vspace{-.5em}}

\author{
 Ilija Radosavovic \quad Piotr Doll\'ar \quad Ross Girshick \quad Georgia Gkioxari \quad Kaiming He \vspace{.5em}\\
 Facebook AI Research (FAIR) \vspace{-1em}
}

\maketitle

\begin{abstract}
\vspace{-.5em}
We investigate \emph{omni-supervised learning}, a special regime of semi-supervised learning in which the learner exploits all available labeled data plus internet-scale sources of unlabeled data. Omni-supervised learning is lower-bounded by performance on existing labeled datasets, offering the potential to surpass state-of-the-art fully supervised methods. To exploit the omni-supervised setting, we propose \emph{data distillation}, a method that ensembles predictions from multiple transformations of unlabeled data, using a single model, to automatically generate new training annotations. We argue that visual recognition models have recently become accurate enough that it is now possible to apply classic ideas about self-training to challenging real-world data. Our experimental results show that in the cases of human keypoint detection and general object detection, state-of-the-art models trained with data distillation surpass the performance of using labeled data from the COCO dataset alone.
\end{abstract}
\vspace{-1em}
\section{Introduction}

This paper investigates \emph{omni-supervised learning}, a paradigm in which the learner exploits as much well-annotated data as possible (\eg, ImageNet \cite{Deng2009}, COCO \cite{Lin2014}) and is also provided with potentially unlimited unlabeled data (\eg, from internet-scale sources). It is a special regime of semi-supervised learning. However, most research on semi-supervised learning has \emph{simulated} labeled/unlabeled data by splitting a fully annotated dataset and is therefore likely to be \emph{upper-bounded} by fully supervised learning with all annotations. On the contrary, omni-supervised learning is \emph{lower-bounded} by the accuracy of training on all annotated data, and its success can be evaluated by how much it surpasses the fully supervised baseline.

\begin{figure}[t]
\centering
\includegraphics[width=1\linewidth]{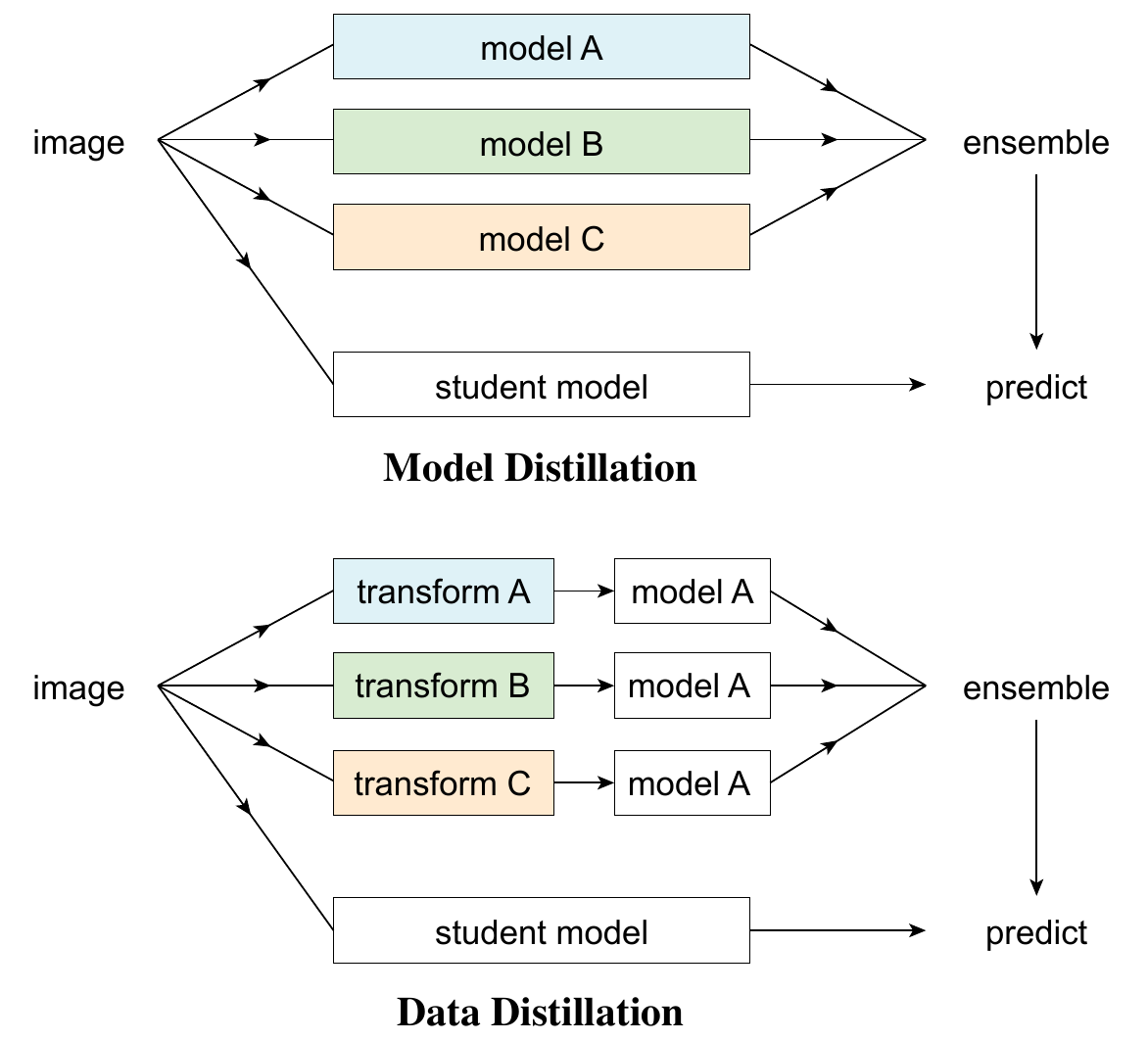}
\caption{\textbf{Model Distillation \cite{Hinton2015} \vs Data Distillation.} In data distillation, ensembled predictions from a single model applied to multiple transformations of an unlabeled image are used as automatically annotated data for training a student model.}
\label{fig:teaser}
\vspace{-.5em}
\end{figure}

To tackle omni-supervised learning,
we propose to perform knowledge distillation \emph{from data}, inspired by \cite{Bucilua2006,Hinton2015} which performed knowledge distillation \emph{from models}.
Our idea is to generate annotations on unlabeled data using a model trained on large amounts of labeled data, and then retrain the model using the extra generated annotations. However, training a model on its own predictions often provides no meaningful information. We address this problem by ensembling the results of a single model run on different transformations (\eg, flipping and scaling) of an unlabeled image. Such transformations are widely known to improve single-model accuracy \cite{Krizhevsky2012} when applied at test time, indicating that they can provide nontrivial knowledge that is not captured by a single prediction. In other words, in comparison with \cite{Hinton2015}, which distills knowledge from the predictions of multiple models, we distill the knowledge of a single model run on multiple transformed copies of unlabeled data (see Figure~\ref{fig:teaser}).

Data distillation is a simple and natural approach based on ``self-training'' (\ie, making predictions on unlabeled data and using them to update the model), related to which there have been continuous efforts \cite{Scudder1965,Yarowsky1995,Vapnik1998,Rosenberg2005,Li2007,Weston2008b,Chen2013,Laine2017} dating back to the 1960s, if not earlier. However, our simple data distillation approach can become realistic largely thanks to the rapid improvement of fully-supervised models \cite{Krizhevsky2012,Simonyan2015,Szegedy2015,He2016,Girshick2014,Girshick2015,Ren2015,Pinheiro2015,Long2015,He2017} in the past few years. In particular, we are now equipped with accurate models that may make fewer errors than correct predictions. This allows us to trust their predictions on unseen data and reduces the requirement for developing data cleaning heuristics. As a result, data distillation does not require one to change the underlying recognition model (\eg, no modification on the loss definitions), and is a scalable solution for processing large-scale unlabeled data sources.

To test data distillation for omni-supervised learning, we evaluate it on the human keypoint detection task of the COCO dataset \cite{Lin2014}. We demonstrate promising signals on this real-world, large-scale application. Specifically, we train a Mask R-CNN model \cite{He2017} using data distillation applied on the original labeled COCO set and another large unlabeled set (\eg, static frames from Sports-1M \cite{Karpathy2014}).
Using the distilled annotations on the unlabeled set, we have observed improvement of accuracy on the held-out validation set: \eg, we show an up to 2 points AP improvement over the strong Mask R-CNN baseline. As a reference, this improvement compares favorably to the $\app$3 points AP improvement gained from training on a similar amount of extra \emph{manually labeled} data in \cite{Papandreou2017} (using private annotations). We further explore our method on COCO object detection and show gains over fully-supervised baselines.

\section{Related Work}

Ensembling \cite{Hansen1990} multiple models has been a successful method for improving accuracy. Model compression \cite{Bucilua2006} is proposed to improve test-time efficiency of ensembling by compressing an ensemble of models into a single student model. This method is extended in knowledge distillation \cite{Hinton2015}, which uses soft predictions as the student's target.

The idea of distillation has been adopted in various scenarios. FitNet \cite{Romero2015} adopts a shallow and wide teacher models to train a deep and thin student model. Cross modal distillation \cite{Gupta2016} is proposed to address the problem of limited labels in a certain modality. In \cite{Lopez2016} distillation is unified with privileged information \cite{Vapnik2015}. To avoid explicitly training multiple models, Laine and Aila \cite{Laine2017} exploit multiple checkpoints during training to generate the ensemble predictions. Following the success of these existing works, our approach distills knowledge from a lightweight ensemble formed by
multiple data transformations.

There is a great volume of work on semi-supervised learning, and comprehensive surveys can be found in \cite{Zhu2005,Chapelle2006,Zhu2009}. Among semi-supervised methods, our method is most related to self-training, a strategy in which a model's predictions on unlabeled data are used to train itself \cite{Scudder1965,Yarowsky1995,Vapnik1998,Rosenberg2005,Li2007,Weston2008b,Chen2013,Laine2017}. 
Closely related to our work on keypoint/object detection, Rosenberg \etal \cite{Rosenberg2005} demonstrate that self-training can be used for training object detectors. Compared to prior efforts, our method is substantially simpler. Once the predicted annotations are generated, our method leverages them as if they were true labels; it does not require any modifications to the optimization problem or model structure.

Multiple views or perturbations of the data can provide useful signal for semi-supervised learning. In the co-training framework \cite{Blum1998}, different views of the data are used to learn two distinct classifiers that are then used to train one another over unlabeled data. Reed \etal \cite{Reed2014} use a reconstruction consistency term for training classification and detection models. Bachman \etal \cite{Bachman2014} employ the pseudo-ensemble regularization term to train models robust on input perturbations. Sajjadi \etal \cite{Sajjadi2016} enforce consistency between outputs computed for different transformations of input examples. 
Simon \etal \cite{Simon2017} utilize multi-view geometry to generate hand keypoint labels from multiple cameras and retrain the detector. 
In an auto-encoder scenario, Hinton \etal \cite{Hinton2006} propose to use multiple ``capsules'' to model multiple geometric transformations.
Our method is also based on multiple geometric transformations, but it does not require to modify network structures or impose consistency by adding any extra loss terms.

Regarding the large-scale regime, Fergus \etal \cite{Fergus2009} investigate semi-supervised learning on 80 millions tiny images. A Never Ending Image Learner (NEIL) \cite{Chen2013} employs self-training to perform semi-supervised learning from web-scale image data. These methods were developed before the recent renaissance of deep learning. In contrast, our method is evaluated with strong deep neural network baselines, and can be applied to structured prediction problems beyond image-level classification (\eg, keypoints and boxes).

\begin{figure*}[t]
\centering
\includegraphics[width=0.95\linewidth]{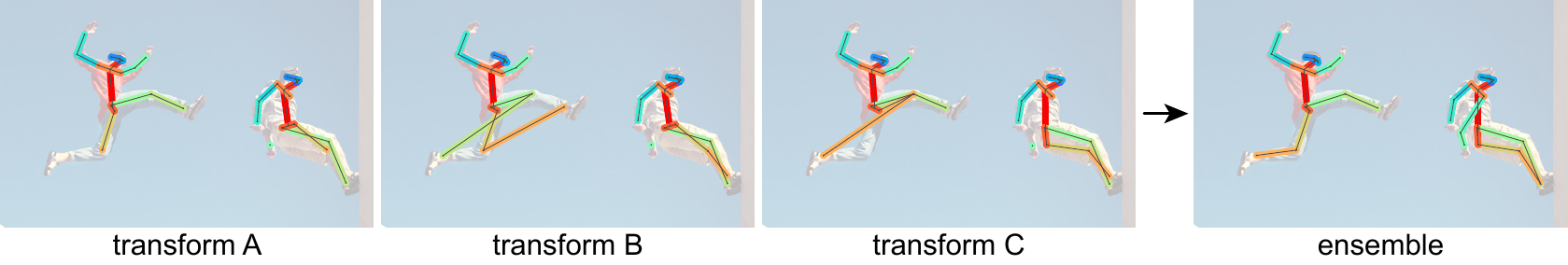}
\caption{\textbf{Ensembling keypoint predictions from multiple data transformations can yield a single superior (automatic) annotation.} For visualization purposes all images and keypoint predictions are transformed back to their original coordinate frame.}
\label{fig:transform_preds}
\vspace{-1em}
\end{figure*}

\section{Data Distillation}

We propose \emph{data distillation}, a general method for omni-supervised learning that distills knowledge from unlabeled data without the requirement of training a large set of models. Data distillation involves four steps: (1) training a model on manually labeled data (just as in normal supervised learning); (2) applying the trained model to multiple transformations of unlabeled data; (3) converting the predictions on the unlabeled data into labels by ensembling the multiple predictions; and (4) retraining the model on the union of the manually labeled data and automatically labeled data. We describe steps 2-4 in more detail below.

\paragraph{Multi-transform inference.} A common strategy for boosting the accuracy of a visual recognition model is to apply the same model to multiple transformations of the input and then to aggregate the results. Examples of this strategy include using multiple crops of an input image (\eg, \cite{Krizhevsky2012,Szegedy2014}) or applying a detection model to multiple image scales and merging the detections (\eg, \cite{Viola2001,Felzenszwalb2010,Dollar2014,Sermanet2014}). We refer to the general application of inference to multiple transformations of a data point with a single model as \emph{multi-transform inference}. In data distillation, we apply multi-transform inference to a potentially massive set of unlabeled data.

\paragraph{Generating labels on unlabeled data.}
By aggregating the results of multi-transform inference, it is often possible to obtain a single prediction that is superior to any of the model's predictions under a single transform (\eg, see Figure \ref{fig:transform_preds}). Our observation is that the aggregated prediction generates \emph{new knowledge} and in principle the model can use this information to learn from itself by generating labels.

Given an unlabeled image and a set of predictions from multi-transform inference, there are multiple ways one could automatically generate labels on the image. For example, in the case of a classification problem the image could be labeled with the average of the class probabilities \cite{Hinton2015}. This strategy, however, has two problems. First, it generates a ``soft'' label (a probability vector, not a categorical label) that may not be straightforward to use when retraining the model. The training loss, for example, may need to be altered such that its compatible with soft labels. Second, for problems with structured output spaces, like object detection or human pose estimation, it does not make sense to average the output as care must be taken to respect the structure of the output space.

Given these considerations, we simply ensemble (or aggregate) the predictions from multi-transform inference in a way that generates ``hard'' labels of the same structure and type of those found in the manually annotated data. Generating hard labels typically requires a small amount of task-specific logic that addresses the structure of the problem (\eg, merging multiple sets of boxes by non-maximum suppression). Once such labels are generated, they can be used to retrain the model in a simple plug-and-play fashion, as if they were authentic ground-truth labels.

Finally, we note that while this procedure requires running inference multiple times, it is actually \emph{efficient} because it is generally substantially less expensive than training multiple models from scratch, as is required by model distillation \cite{Bucilua2006,Hinton2015}.

\paragraph{Knowledge distillation.} The new knowledge generated from unlabeled data can be used to improve the model. To do this, a student model (which can be the same as the original model or different) is trained on the \emph{union} set of the original supervised data and the unlabeled data with automatically generated labels.

Training on the union set is straightforward and requires no change to the loss function. However, we do take two factors into considerations. First, we ensure that each training minibatch contains a mixture of manually labeled data and automatically labeled data. This ensures that every minibatch has a certain percentage of ground-truth labels, which results in better gradient estimates. Second, since more data is available, the training schedule must be lengthened to take full advantage of it. We discuss these issues in more detail in the context of the experiments.

\begin{figure*}[t]
\begin{center}
\includegraphics[width=1.0\linewidth]{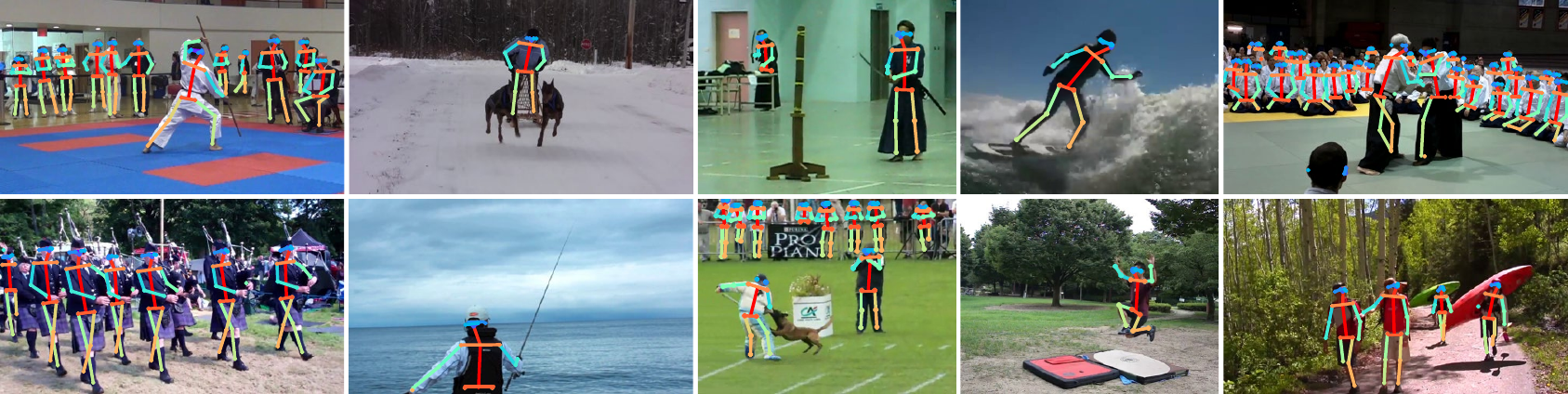}
\end{center}
\vspace{-1em}
\caption{\textbf{Random examples of annotations generated on static Sports-1M \cite{Karpathy2014} frames using a ResNet-50 teacher.} The generated annotations have reasonably high quality, though as expected there are mistakes like inverted keypoints (top right).}
\label{fig:s1m_examples}
\vspace{-.5em}
\end{figure*}

\section{Data Distillation for Keypoint Detection} \label{sec:kp_detection}

This section describes an instantiation of data distillation for the application of multi-person keypoint detection.

\paragraph{Mask R-CNN.} Our teacher and student models are the Mask R-CNN \cite{He2017} keypoint detection variant. Mask R-CNN is a two-stage model. The first stage is a Region Proposal Network (RPN) \cite{Ren2015}. The second stage consists of three heads for bounding box classification, regression, and keypoint prediction on each Region of Interest (RoI). The keypoint head outputs a heatmap that is trained to predict a one-hot mask for each keypoint type.
We use ResNet \cite{He2016} and ResNeXt \cite{Xie2017} with Feature Pyramid Networks (FPN) \cite{Lin2017} as backbones for Mask R-CNN. All implementations follow \cite{He2017}, unless specified.

\paragraph{Data transformations.} This paper opts for \emph{geometric} transformations for multi-transform inference, though other transformations such as color jittering \cite{Krizhevsky2012} are possible. The only requirement is that it must be possible to ensemble the resulting predictions. For geometric transformations, if the prediction is a geometric quantity (\eg, coordinates of a keypoint), then the inverse transformation must be applied to each prediction before they are merged.

 We use two popular transformations: scaling and horizontal flipping. We resize the unlabeled image to a pre-defined set of scales (denoted by the shorter side of an image): [400, 1200] pixels with a stepsize of 100, which was selected by measuring the keypoint AP for the teacher model when applying these transformations on the validation set. The selected transformations can improve the model by a good margin, \eg for ResNet-50 from 65.1 to 67.8 AP, which is then used as the teacher.
Note that unless stated, we do \emph{not} apply these transformation at test time for all baseline/distilled models.

Figure~\ref{fig:s1m_examples} shows some examples of the generated annotations on Sport-1M. They have reasonably high quality.

\begin{figure*}[t]
\begin{center}
\includegraphics[width=1.0\linewidth]{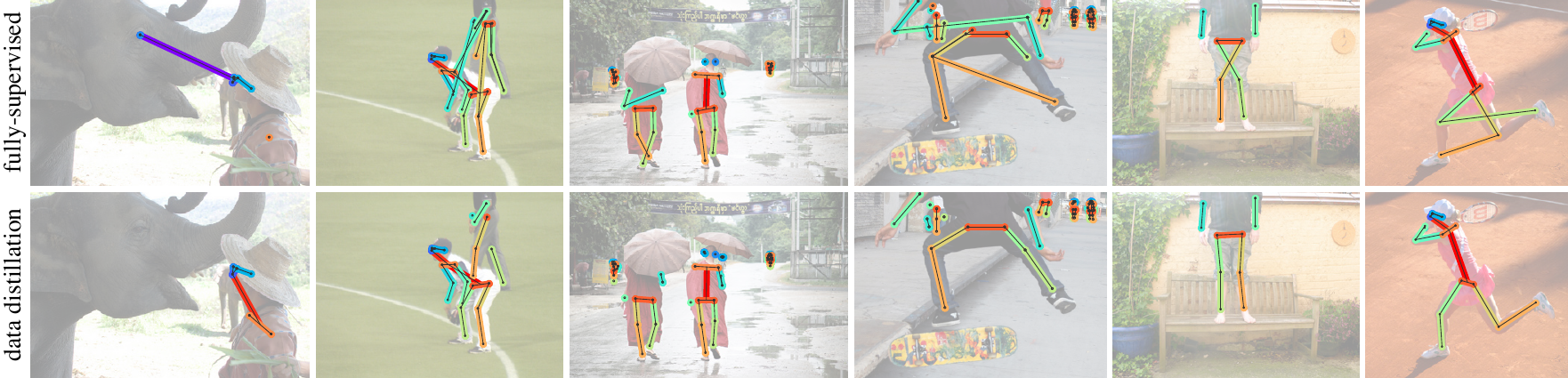}
\end{center}
\vspace{-1em}
\caption{\textbf{Selected results of fully-supervised learning in the original \texttt{co-115} set (top) \vs data distillation in the \texttt{co-115} plus \texttt{s1m-180} sets (bottom).} The results are on the held-out data from COCO \texttt{test-dev}.}
\label{fig:w_vs_wo_kdd}
\vspace{-.5em}
\end{figure*}

\paragraph{Ensembling.} One could ensemble the multi-transform inference results from each stage and each head of Mask R-CNN. In our experiments, however, for simplicity we only apply multi-transform inference to the keypoint head; the outputs from the other stage (\ie, RPN) and heads (\ie, bounding box classification and regression) are from a single-scale without any transformations.

Thanks to the above simplification, it is easy for us to have a consistent set of detection boxes serving as the RoIs for all transformations (scales/flipping). On a single RoI, we extract the keypoint heatmaps from all transformations, and although they are from different geometric transformations, these heatmaps are with reference to the local coordinate system of the same RoI. So we can directly average the output (probability) of these heatmaps for ensembling. We take the argmax position in this ensembling result and generate the predicted keypoint location.

\paragraph{Selecting predictions.} We expect the predicted boxes and keypoints to be reliable enough for generating good training labels. Nevertheless, the predictions will contain \emph{false positives} that we hope to identify and discard. We use the predicted detection score as a proxy for prediction quality and generate annotations only from the predictions that are above a certain score threshold.
In practice, we found that a score threshold works well if it makes ``the average number of annotated instances per unlabeled image'' roughly equal to ``the average number of instances per labeled image''. Although this heuristic assumes that the unlabeled and labeled images follow similar distributions, we found that it is robust and works well even in cases where the assumption does not hold.

As a dual consideration to false positives above, there may be \emph{false negatives} (\ie, missing detections) in the extra data, and the annotations generated should not necessarily be viewed as \emph{complete} (\ie, absence of an annotation does not imply true background). However, in our practice we have tried either to sample or not sample background regions from the extra data for training detectors, and have observed no difference in accuracy. For simplicity, in all experiments we view the generated data as complete, so the extra data are simply treated as if all annotations are correct.

\paragraph{Generating keypoint annotations.} Each of the selected predictions consists of $K$ individual keypoints (\eg, left ear, nose, \etc). Since many of the object views do not show all of the keypoint types, the predicted keypoints are likely to contain false positives as well.
As above, we choose a threshold such that the average numbers of keypoints are approximately equal in the supervised and generated sets.

\paragraph{Retraining.}
We train a student model on the union set of the original supervised images and the images with automatically generated annotations. To maintain supervision quality at the minibatch level, we use a fixed sampling ratio for the two kinds of data. Specifically, we randomly sample images for each minibatch such that the expected ratio of original images to generated labeled images is 6:4, unless stated otherwise.

We adopt the learning rate schedule similar to \cite{He2017} and increase the total number of iterations to account for extra images. The learning rate starts from $0.02$ and is divided by $10$ after $70\%$ and $90\%$ of the total number of iterations. The impact of the total number of iterations will be discussed in the next section in context of Table~\ref{tab:ablations:num_iter}.

We use a student model with the same architecture as the teacher. The student can either be fine-tuned starting from the teacher model or retrained from the initial weights (\ie, those pre-trained on ImageNet \cite{Russakovsky2015}). We found that \emph{retraining} consistently results in better performance, suggesting that the teacher model could have been in a poor local optimum. We opt for retraining in all experiments.

\section{Experiments on Keypoint Detection}

We evaluate data distillation on the keypoint detection task of the COCO dataset \cite{Lin2014}. We report keypoint Average Precision following the COCO definitions, including AP (COCO's default, averaged over different IoU thresholds), AP$_{50}$, AP$_{75}$, AP$_{M}$ (medium), and AP$_L$ (large). In all experiments we report results on the 2017 validation set that contains 5k images (called \texttt{val2017}, formerly known as \texttt{minival}). 

\subsection{Data Splits}

Our experiments involve several splits of data:

\paragraph{COCO labeled images.} These are the original labeled COCO images that contain ground-truth person and keypoint annotations. 
In this paper, we refer to the 80k training images as \bd{\texttt{co-80}}, a 35k subset of the 2014 validation images as \bd{\texttt{co-35}}, and their union as \bd{\texttt{co-115}} (in the 2017 version of COCO, \texttt{co-115} is the \texttt{train2017} set). We do not use the original train/val nomenclature because their roles may change in different experiments.

\paragraph{COCO unlabeled images.} The 2017 version of COCO provides a collection of 120k unlabeled images, which we call \bd{\texttt{un-120}}. These images are expected to have a \emph{similar distribution} as the labeled COCO images.

\paragraph{Sports-1M static frames.} We will show that our method can be robust to a \emph{dissimilar distribution} of unlabeled data.
We collect these images by using \emph{static frames} from the Sports-1M \cite{Karpathy2014} video dataset. 
We randomly sample 180k videos from this dataset. Then we randomly sample 1 frame from each video, noting that we do \emph{not} exploit any temporal information even if it is possible. This strategy gives us 180k static images. We call this set \bd{\texttt{s1m-180}}. We do not use any available labels from this static image set. 

\subsection{Main Results}

We investigate data distillation in three cases:

\noindent \textbf{(i)} Small-scale data as a sanity check: we use \texttt{co-35} as the labeled data and treat \texttt{co-80} as unlabeled.

\noindent \textbf{(ii)} Large-scale data with similar distribution: we use \texttt{co-115} as the labeled data and \texttt{un-120} as unlabeled.

\noindent \textbf{(iii)} Large-scale data with dissimilar distribution: we use \texttt{co-115} as the labeled data and \texttt{s1m-180} as unlabeled.

\noindent The results are in Table~\ref{tab:regimes}, discussed as follows:

\paragraph{Small-scale data.} As a sanity-check, we evaluate our approach in the classic semi-supervised setting by simulating labeled and unlabeled splits from all labeled images.

In Table~\ref{tab:regime:small_kdd}, we show results of data distillation performed on \texttt{co-35} as the labeled data and \texttt{co-80} treated as unlabeled data. As a comparison, we report supervised learning results using either \texttt{co-35} or \texttt{co-115}. This comparison shows that data distillation is a successful semi-supervised learning method: it surpasses the \texttt{co-35}-only counterpart by 5.3 points of AP by using unlabeled data (60.2 \vs 54.9). On the other hand, as expected, the semi-supervised learning result is lower than fully-supervised learning on \texttt{co-115} (60.2 \vs 65.1).

This phenomenon on small-scale data has been widely observed for many semi-supervised learning methods and datasets: if labels \emph{were} available for all training data, then the accuracy of semi-supervised learning would be \emph{upper-bounded} by using all labels. In addition, as the \emph{simulated} splits are often at smaller scales, there is a relatively large gap for the semi-supervised method to improve in (\eg, from 54.9 to 65.1).

We argue that omni-supervised learning is a real-world scenario unlike the above simulated semi-supervised setting.
Even though one could label many images, there are always more unlabeled data available (\eg, at internet-scale). We can thus pursue an accuracy that is \emph{lower-bounded}.
In addition, when trained with a larger dataset, the supervised baseline would be much \emph{higher} (\eg, 65.1), leaving \emph{less room} for models to gain from the unlabeled data.

Therefore, we argue that the large-scale, high-accuracy regime is more challenging and of more interest in practice. We investigate it in the following experiments.

\begin{table}[t]
\subfloat[\textbf{Small-scale data.}  Data distillation is performed on \texttt{co-35} with labels and \texttt{co-80} without labels, \vs fully-supervised learning performed on \texttt{co-35} and \texttt{co-115} respectively. The backbone is ResNet-50.\label{tab:regime:small_kdd}]{
\makebox[1\linewidth]{
\tablestyle{4pt}{1.05}
\begin{tabular}{cc|x{20}x{20}x{20}x{20}x{20}}
labeled              & unlabeled             &   AP          &  AP$_{50}$  & AP$_{75}$   & AP$_{M}$    & AP$_{L}$    \\
 \shline
\texttt{co-35}  &                 &          54.9 &        80.5 &        59.0 &        50.1 &        62.8 \\
\texttt{co-35}  & \texttt{co-80}  &   \bftab 60.2 & \bftab 83.8 & \bftab 65.4 & \bftab 55.2 & \bftab 68.4 \\
\texttt{co-115} &                 &          \demph{65.1} &        \demph{86.6} &        \demph{70.9} &        \demph{59.9} &        \demph{73.6}
\end{tabular}
}}\\
\subfloat[\textbf{Large-scale, similar-distribution data.} Data distillation (DD) is performed on \texttt{co-115} with labels and \texttt{un-120} without labels, comparing with the supervised counterparts trained on \texttt{co-115}.\label{tab:regime:large_coco_kdd}]{
\makebox[1\linewidth]{
\tablestyle{2.5pt}{1.05}
\begin{tabular}{l|x{15}|x{20}x{20}x{20}x{20}x{20}}
 backbone                    &  DD        & AP           &  AP$_{50}$  & AP$_{75}$   & AP$_{M}$    & AP$_{L}$    \\
 \shline
 ResNet-50                   &            &        65.1  &        86.6 &        70.9 &        59.9 &        73.6 \\
 ResNet-50                   & \checkmark & \bftab 67.1  & \bftab 87.9 & \bftab 73.4 & \bftab 62.2 & \bftab 75.1 \\
 \hline
 ResNet-101                  &            &        66.1  &        87.7 &        71.7 &        60.5 &        75.0 \\
 ResNet-101                  & \checkmark & \bftab 67.8  & \bftab 88.2 & \bftab 73.8 & \bftab 62.8 & \bftab 76.0 \\
 \hline
 ResNeXt-101-32$\times$4     &            &        66.8  &        87.5 &        73.0 &        61.6 &        75.2 \\
 ResNeXt-101-32$\times$4     & \checkmark & \bftab 68.7  & \bftab 88.9 & \bftab 75.1 & \bftab 63.9 & \bftab 76.7 \\
 \hline
 ResNeXt-101-64$\times$4     &            &        67.3  &        88.0 &        73.3 &        62.2 &        75.6 \\
 ResNeXt-101-64$\times$4     & \checkmark & \bftab 69.1  & \bftab 88.9 & \bftab 75.3 & \bftab 64.1 & \bftab 77.1
\end{tabular}
}}\\
\subfloat[\textbf{Large-scale, dissimilar-distribution data.} Data distillation (DD) is performed on \texttt{co-115} with labels and \texttt{s1m-180} without labels, comparing with the supervised counterparts trained on \texttt{co-115}.\label{tab:regime:large_s1m_kdd}]{
\makebox[1\linewidth]{
\tablestyle{2.5pt}{1.05}
\begin{tabular}{l|x{15}|x{20}x{20}x{20}x{20}x{20}x{20}}
 backbone                    & DD         & AP           &  AP$_{50}$  & AP$_{75}$   & AP$_{M}$    & AP$_{L}$    \\
 \shline
 ResNet-50                   &            &        65.1  &        86.6 &        70.9 &        59.9 &        73.6 \\
 ResNet-50                   & \checkmark & \bftab 66.6  & \bftab 87.3 & \bftab 72.6 & \bftab 61.6 & \bftab 75.0 \\
 \hline
 ResNet-101                  &            &        66.1  &        87.7 &        71.7 &        60.5 &        75.0 \\
 ResNet-101                  & \checkmark & \bftab 67.5  & \bftab 87.9 & \bftab 73.9 & \bftab 62.4 & \bftab 75.9 \\
 \hline
 ResNeXt-101-32$\times$4     &            &        66.8  &        87.5 &        73.0 &        61.6 &        75.2 \\
 ResNeXt-101-32$\times$4     & \checkmark & \bftab 68.0  & \bftab 88.1 & \bftab 74.2 & \bftab 63.1 & \bftab 76.2 \\
 \hline
 ResNeXt-101-64$\times$4     &            &        67.3  &        88.0 &        73.3 &        62.2 &        75.6 \\
 ResNeXt-101-64$\times$4     & \checkmark & \bftab 68.5  & \bftab 88.8 & \bftab 74.9 & \bftab 63.7 & \bftab 76.5
\end{tabular}
}}\vspace{.5em}
\caption{Data distillation for COCO keypoint detection. Keypoint AP is reported on COCO \texttt{val2017}.}
\label{tab:regimes}\vspace{-.5em}
\end{table}

\paragraph{Large-scale, similar-distribution data.} Table~\ref{tab:regime:large_coco_kdd} shows the scenario of a real-world omni-supervised learning application: we have a large-scale source of 120k COCO (\texttt{un-120}) images on hand, but \emph{we do not have labels for them}. Can we improve over our best baseline results using these unlabeled data?

Table~\ref{tab:regime:large_coco_kdd} shows the data distillation results on \texttt{co-115} plus \texttt{un-120}, comparing with the fully-supervised counterpart on \texttt{co-115}, the largest available annotated set on hand. Our method is able to improve over the strong baselines by 1.7 to 2.0 points AP. Our improvement is observed regardless of the depth/capacity of the backbone models, including ResNet-50/101 and ResNeXt-101.

We argue that these are non-trivial results. Because the baselines are very high due to using large amounts of supervised data (115k images in \texttt{co-115}), they might leave less room for further improvement, in contrast to the simulated semi-supervised setting. Actually, in recent work \cite{Papandreou2017} that exploited an \emph{extra} 1.5$\times$ \emph{fully-annotated} human keypoint skeletons (contributed by in-house annotators), the improvement is $\app$3 points AP over their baseline. Given this context, our increase of $\app$2 points AP, contributed by a similar amount of extra \emph{unlabeled} data, is very promising.

\paragraph{Large-scale, dissimilar-distribution data.} Even though COCO data are images ``in the wild'', the \texttt{co-115} and \texttt{un-120} sets are subject to similar data distributions. As one further step toward omni-supervision in real cases, we investigate a scenario where the unlabeled images are from a \emph{different distribution}.

Table~\ref{tab:regime:large_s1m_kdd} shows data distillation results on \texttt{co-115} plus \texttt{s1m-180}. Comparing with the supervised baselines trained on \texttt{co-115}, our method shows consistent improvement with different backbones, achieving 1.2 to 1.5 points of AP increase. Moreover, the improvements in this case are reasonably close to those in Table~\ref{tab:regime:large_coco_kdd}, even though the data distribution in Sport-1M is different. This experiment shows that our method, in the application of keypoint detection, is robust to the misaligned distribution of data.
This is a promising signal for real-world omni-supervised learning. 

Figure~\ref{fig:w_vs_wo_kdd} shows some examples of the fully-supervised results trained in \texttt{co-115} and the data distillation results trained in \texttt{co-115} plus \texttt{s1m-180}.

\subsection{Ablation Experiments}

In addition to the above main results, we conduct several ablation experiments as analyzed in the following:

\begin{table}[t]
\makebox[1\linewidth]{
\tablestyle{4pt}{1.05}
\begin{tabular}{rr|x{20}x{20}x{20}x{20}x{20}}
                           & \#iter & AP           &  AP$_{50}$  & AP$_{75}$   & AP$_{M}$    & AP$_{L}$    \\
 \shline
 \emph{fully-supervised}   & 90k    &         64.2 &        86.4 &        69.2 &        59.1 &        72.6 \\
                           & 130k   &    \bd{65.1} &   \bd{86.6} &   \bd{70.9} &   \bd{59.9} &   \bd{73.6} \\
                           & 270k   &         64.7 &   \bd{86.6} &        70.4 &        59.7 &        73.0 \\
 \hline
  \emph{data distillation} & 90k    &         63.6 &        85.9 &        69.2 &        58.8 &        71.7 \\
                           & 180k   &         65.8 &        87.3 &        71.6 &        60.8 &        74.2 \\
                           & 270k   &         66.5 &   \bd{88.0} &        72.2 &        61.5 &        74.6 \\
                           & 360k   &    \bd{66.6} &       87.3  &   \bd{72.6} &   \bd{61.6} &   \bd{75.0}
\end{tabular}
}
\vspace{.3em}
\caption{Ablation on \textbf{numbers of training iterations.} The models are trained on \texttt{co-115} (and plus \texttt{s1m-180} for data distillation). The backbone is ResNet-50. In all case, the learning rate is reduced by 10 at $70\%$ and $90\%$ of the total number of iterations.}
\label{tab:ablations:num_iter}
\vspace{-.5em}
\end{table}

\paragraph{Number of iterations.} It is necessary to train for more iterations when given more (labeled or distilled) data. To show that our method does \bd{\emph{not}} simply take advantage of longer training, we conduct a careful ablation experiment on the number of iterations in Table~\ref{tab:ablations:num_iter}.

For the fully-supervised baseline, we investigated a total number of iterations of 90k (as done in \cite{He2017}), 130k ($\app$1.5$\times$ longer), and 270k (3$\times$ longer). Table~\ref{tab:ablations:num_iter} (top) shows that an appropriately long training indeed leads to better results, and the original schedule of 90k in \cite{He2017} is suboptimal. However, without increasing the dataset size, training longer gives diminishing return and becomes \emph{prone to overfitting}. The optimal number of 130k iterations is chosen and used in Tables~\ref{tab:regimes} for the fully-supervised baselines.

In contrast, our data distillation method continuously improves when the number of iterations is increased from 90k to 360k as shown in Table~\ref{tab:ablations:num_iter} (bottom). With a short training of 90k, our method is inferior to its fully-supervised counterpart (63.6 \vs 64.2), which is understandable: the generated labels in the extra data have lower quality than the ground-truth labels, and the model may not benefit from them unless ground-truth labels have been sufficiently exploited. On the other hand, our method starts to show a healthy gain with sufficient training and surpasses its fully-supervised counterpart. Actually, our method's performance has not saturated and is likely to improve when using more iterations. To have manageable experiments, for all other data distillation results in the paper, our method uses 360k iterations.

\begin{figure}[t]
\centering
\includegraphics[width=0.75\linewidth]{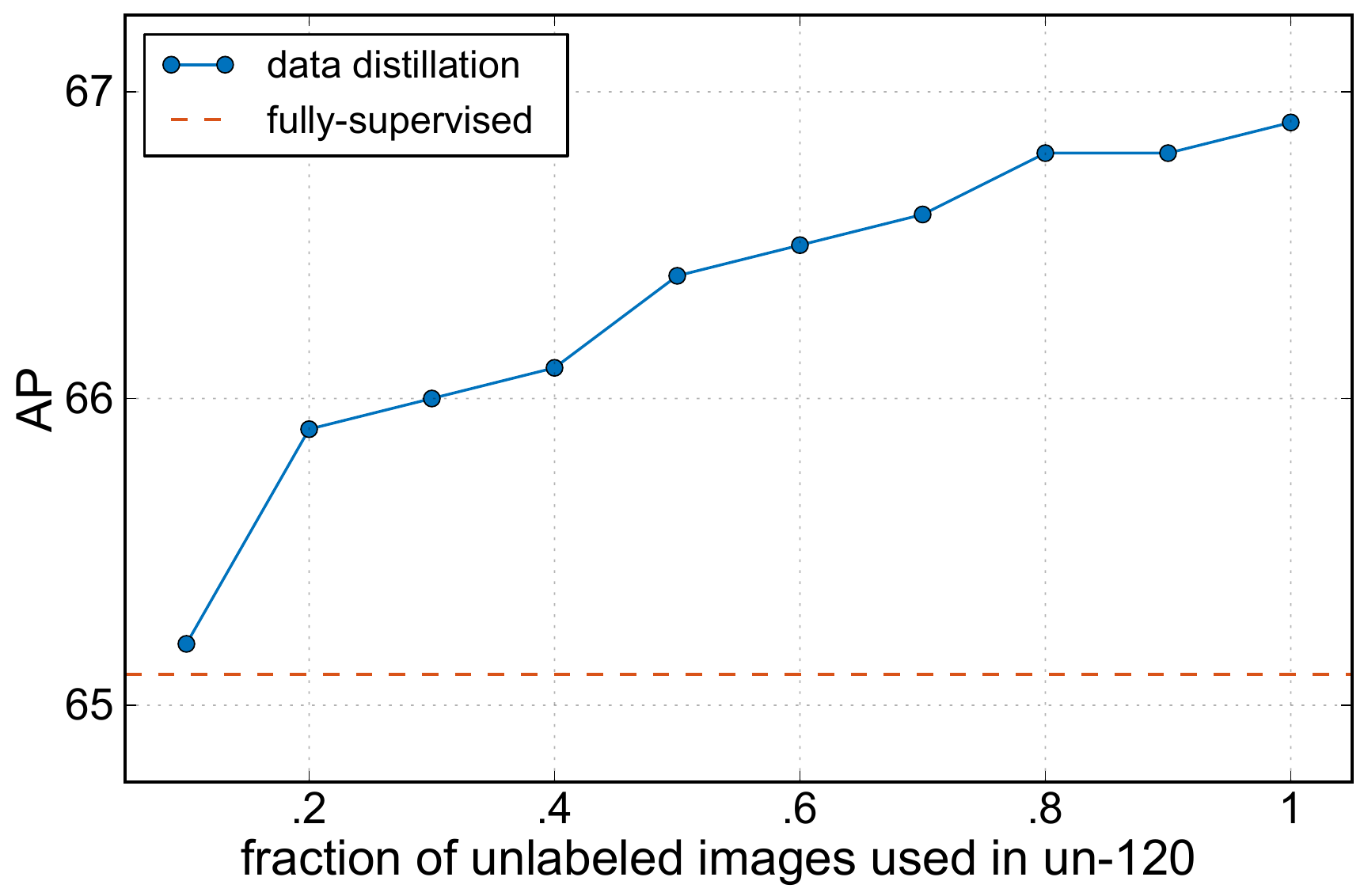}\quad\quad
\caption{Data distillation applied to \texttt{co-115} with labels and \textbf{different fractions of \texttt{un-120} images without labels}, comparing with the \texttt{co-115} fully-supervised baseline, using ResNet-50.
}
\vspace{-.3em}
\label{fig:cu_fractions}
\end{figure}

\begin{figure}[t]
\centering
\includegraphics[width=0.75\linewidth]{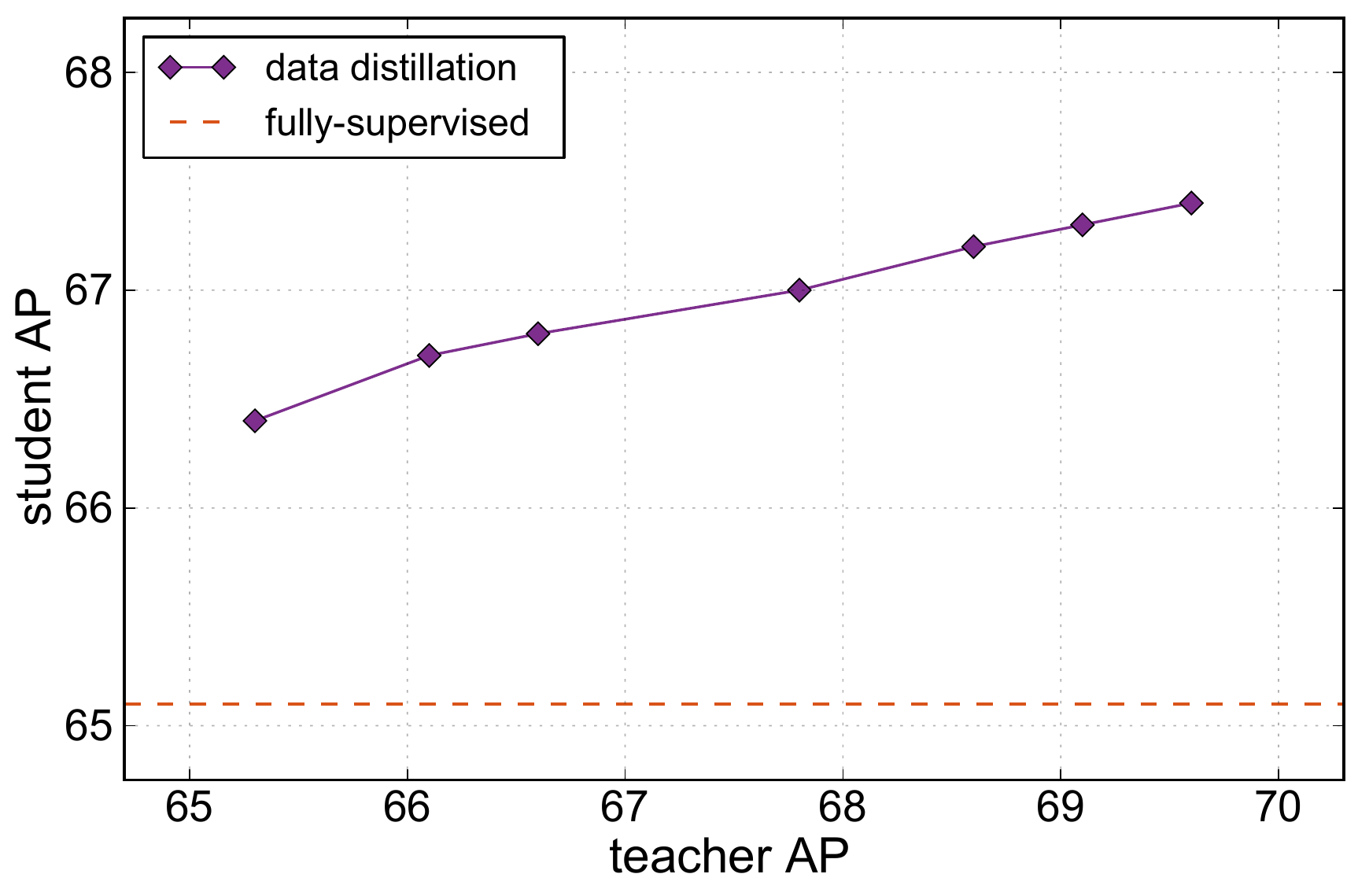}\quad
\vspace{-.3em}
\caption{\textbf{Impact of teacher quality on data distillation.} Student ResNet-50 model is trained using data distillation on \texttt{co-115} plus \texttt{un-120} and compared to its fully-supervised counterpart trained on \texttt{co-115}. Each data point is from a different teacher whose AP can vary because of different backbones, number of supervised training iterations, and transform settings (\eg, scales used).
}
\label{fig:teacher_ablation}
\vspace{-.5em}
\end{figure}

\paragraph{Amount of unlabeled data.} To better understand the importance of the amount of unlabeled data, in Figure~\ref{fig:cu_fractions} we investigate using a subset of the \texttt{un-120} unlabeled data for data distillation (the labeled data is \texttt{co-115}).

To have a simpler unified rule for handling the various sizes of the unlabeled set, for this ablation, we adopt a minibatching and iteration strategy different from the above sections.
Given a fraction $\rho$ of \texttt{un-120} images used, we sample each minibatch with on average 1:$\rho$ examples from the labeled and unlabeled data. The iteration number is adaptively set as 1+$\rho$ times of that of the supervised baseline (130k in this figure).
As such, the total number of sampled images from the labeled set is roughly the same regardless of the fraction $\rho$. We note that this strategy is \emph{suboptimal} comparing with the setting in other tables, but it is a simplified setting that can apply to all fractions investigated.

Figure~\ref{fig:cu_fractions} shows that for all fractions of unlabeled data, our method is able to improve over the supervised baseline. Actually, as can be expected, the supervised baseline becomes a \emph{lower-bound} of accuracy in omni-supervised learning: the extra \emph{unlabeled} data, when exploited appropriately such as in data distillation, should always provide extra information. Moreover, Figure~\ref{fig:cu_fractions} shows that there is a general trend of better results when using more unlabeled data. A similar trend, in the context of \emph{fully-annotated} data, has been observed recently in \cite{Sun2017}. However, our trend is observed in \emph{unlabeled} data and can be more encouraging for the future study in computer vision.

\paragraph{Impact of teacher quality.} To understand the impact of the teacher quality on data distillation, we produce different teacher models with different AP (see Figure~\ref{fig:teacher_ablation} caption). Then we train the same student model on each teacher. Figure~\ref{fig:teacher_ablation} shows the student AP \vs the teacher AP.

As expected, all student models trained by data distillation surpass the fully-supervised baseline. In addition, a higher-quality teacher in general results in a better student. This demonstrates a nice property of the data distillation method: one could expect a bigger improvement if a better teacher will be developed.

\begin{table}[t]
\makebox[1\linewidth]{
\tablestyle{2.5pt}{1.05}
\begin{tabular}{l|c|x{20}x{20}x{20}x{20}x{20}}
 backbone                 & \emph{test aug.?}  &   AP          &  AP$_{50}$  & AP$_{75}$   & AP$_{M}$    & AP$_{L}$    \\
 \shline
 ResNet-50                &                    &          67.1 &        87.9 &        73.4 &        62.2 &        75.1 \\
 ResNet-50                & \checkmark         &   \bftab 68.9 & \bftab 88.8 & \bftab 75.8 & \bftab 64.4 & \bftab 76.4 \\
\hline
 ResNet-101               &                    &          67.8 &        88.2 &        73.8 &        62.8 &        76.0 \\
 ResNet-101               & \checkmark         &   \bftab 69.8 & \bftab 89.1 & \bftab 76.0 & \bftab 65.3 & \bftab 77.5 \\
\hline
 ResNeXt-101-32$\times$4  &                    &          68.7 &        88.9 &        75.1 &        63.9 &        76.7 \\
 ResNeXt-101-32$\times$4  & \checkmark         &   \bftab 70.6 & \bftab 89.3 & \bftab 77.2 & \bftab 65.7 & \bftab 78.4 \\
\hline
 ResNeXt-101-64$\times$4  &                    &          69.1 &        88.9 &        75.3 &        64.1 &        77.1 \\
 ResNeXt-101-64$\times$4  & \checkmark         &   \bftab 70.4 & \bftab 89.3 & \bftab 76.8 & \bftab 65.8 & \bftab 78.1
\end{tabular}
}\vspace{.5em}
\caption{Ablation on \textbf{test-time augmentation.} A data distillation model is trained on \texttt{co-115} + \texttt{un-120}, tested without and with test-time augmentations.}\vspace{-.5em}
\label{tab:ablations:dd_test_aug}
\end{table}

\paragraph{Test-time augmentations.} Our data distillation method exploits multi-transform inference to generate labels. Multi-transform inference can also be applied at test-time to further improve results, a strategy typically called test-time augmentation. Table~\ref{tab:ablations:dd_test_aug} shows the results of applying test-time augmentations on a data distillation model. The augmentations are the same as those used to generate distillation labels. It shows that test-time augmentations can still improve the results over our data distillation model.

Interestingly, the student model's 68.9 AP (ResNet-50, in Table~\ref{tab:ablations:dd_test_aug}) is  higher than its corresponding (test-time augmented) teacher's 67.8 AP.
We believe that this is \emph{a signal of our approach being able to learn new knowledge from the extra unlabeled data}, instead of simply learning to be robust to the transforms. 
Even though we use multiple data-agnostic transforms, the distilled labels are \emph{data-dependent} and may convey knowledge from the extra data.

This result also suggests that performing data distillation in an iterative fashion may improve the results further. We leave this direction for future work.

\begin{figure}[t]
\begin{center}
\includegraphics[width=0.90\linewidth]{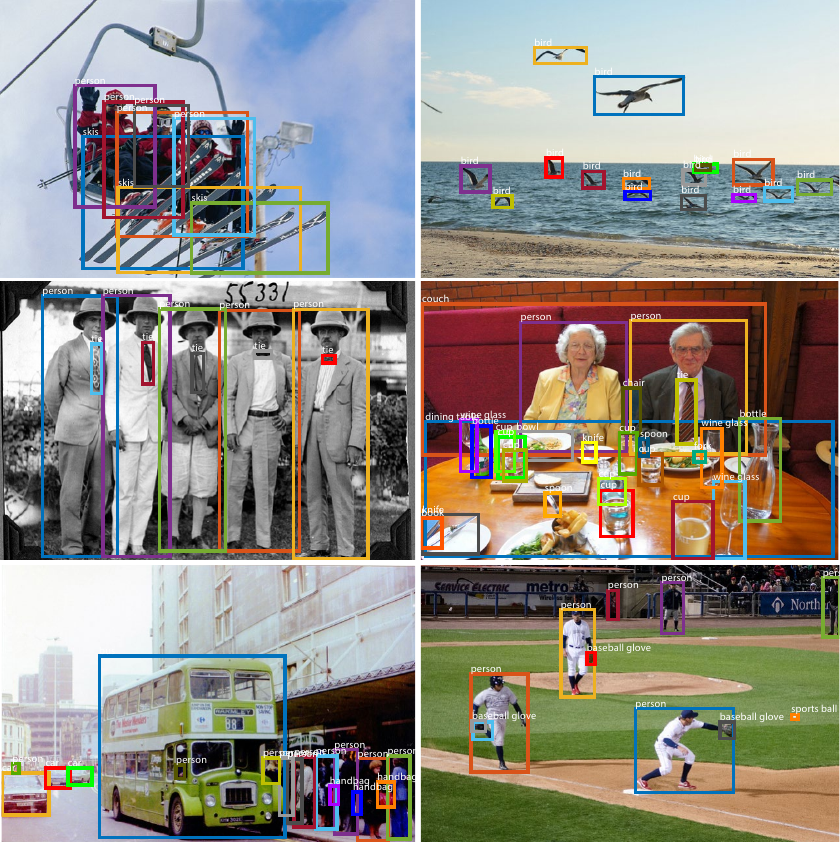}
\end{center}
\vspace{-1em}
\caption{\textbf{Object detection annotations generated on \texttt{un-120}.}}
\label{fig:cu_box_examples}
\vspace{-1em}
\end{figure}

\section{Experiments on Object Detection}

We investigate the generality of our approach by applying it to another task with minimal modification. We perform data distillation for object detection on the COCO dataset \cite{Lin2014}. Here our data splits involve \texttt{co-35/80/115} as defined above. We test on \texttt{minival}.

\subsection{Implementation}
Our object detector is Faster R-CNN \cite{Ren2015} with the FPN backbone \cite{Lin2017} and the RoIAlign improvement \cite{He2017}. We adopt the joint end-to-end training as described in \cite{Ren2017}.
Note that this is unlike in our keypoint experiments where we froze the RPN stage (which created the same set of boxes for keypoint ensembling). To produce the ensemble results, we simply take the union set of the boxes predicted under different transformations, and combine them using bounding box voting \cite{Gidaris2015} (a process similar to non-maximum suppression that merges the suppressed boxes).
This ensembling strategy on the union set of boxes shows the flexibility of our method: it is agnostic to how the results from multiple transformations are aggregated.

The object detection task involves multiple categories.
A single threshold of score for generating labels may lead to strong biases. To address this issue, we set a per-category threshold of score confidence for annotating objects in the unlabeled data. We choose a threshold for each category such that its average number of annotated instances per image in the unlabeled dataset matches the average number of instances in the labeled dataset. Figure~\ref{fig:cu_box_examples} shows some examples of the generated annotations on \texttt{un-120}.

\begin{table}[t]
\subfloat[\textbf{Small-scale data.}  Data distillation is performed on \texttt{co-35} with labels and \texttt{co-80} without labels, \vs fully-supervised learning performed on \texttt{co-35} and \texttt{co-115}. The backbone is ResNet-50.\label{tab:bdd_regime:small_coco}]{
\makebox[1\linewidth]{
\tablestyle{3.4pt}{1.05}
\begin{tabular}{cc|x{20}x{20}x{20}x{20}x{20}x{20}}
 labeled           & unlabeled       & AP           &  AP$_{50}$   & AP$_{75}$    & AP$_{S}$     & AP$_{M}$     & AP$_{L}$     \\
 \shline
 \texttt{co-35}    &                 &        30.5  &        51.9  &        31.9  &        15.2  &        33.0  &        40.6  \\
 \texttt{co-35}    & \texttt{co-80}  & \bftab 32.3  & \bftab 53.8  & \bftab 33.9  & \bftab 16.8  & \bftab 35.5  & \bftab 43.7  \\
 \texttt{co-115}   &                 & \demph{37.1} & \demph{59.1} & \demph{39.6} & \demph{20.0} & \demph{40.0} & \demph{49.4}
\end{tabular}
}}\vspace{-.5em}\\
\subfloat[\textbf{Large-scale data.} Data distillation (DD) is performed on \texttt{co-115} with labels and \texttt{un-120} without labels, comparing with the supervised counterparts trained on \texttt{co-115}.\label{tab:bdd_regime:large_coco}]{
\makebox[1\linewidth]{
\tablestyle{2.2pt}{1.05}
\begin{tabular}{l|x{15}|x{20}x{20}x{20}x{20}x{20}x{20}x{20}}
 backbone                     & DD          &   AP          &  AP$_{50}$  & AP$_{75}$   & AP$_{S}$    & AP$_{M}$    & AP$_{L}$    \\
 \shline
 ResNet-50                    &             &         37.1  &        59.1 &        39.6 &        20.0 &        40.0 &        49.4 \\
 ResNet-50                    & \checkmark  & \bftab  37.9  & \bftab 60.1 & \bftab 40.8 & \bftab 20.3 & \bftab 41.6 & \bftab 50.8 \\
 \hline
 ResNet-101                   &             &         39.2  &        61.0 &        42.3 &        21.7 &        42.9 &        52.3 \\
 ResNet-101                   & \checkmark  & \bftab  40.1  & \bftab 62.1 & \bftab 43.5 & \bftab 21.7 & \bftab 44.3 & \bftab 53.7 \\
 \hline
 ResNeXt-101-32$\times$4      &             &         40.1  &        62.4 &        43.2 &        22.6 &        43.7 &        53.7 \\
 ResNeXt-101-32$\times$4      & \checkmark  & \bftab  41.0  & \bftab 63.3 & \bftab 44.4 & \bftab 22.9 & \bftab 45.5 & \bftab 54.8 \\
\end{tabular}
}}
\vspace{.3em}
\caption{Data distillation for COCO object detection. Box AP is reported on COCO \texttt{val2017}.}
\label{tab:bdd_regimes}
\vspace{-3mm}
\end{table}

\subsection{Object Detection Results} 

We investigate data distillation in two cases (Table~\ref{tab:bdd_regimes}):

\noindent \textbf{(i)} Small-scale data: we use \texttt{co-35} as the labeled data and treat \texttt{co-80} as unlabeled.

\noindent \textbf{(ii)} Large-scale data: we use \texttt{co-115} as the labeled data and \texttt{un-120} as unlabeled.

\paragraph{Small-scale data.} Similar to the keypoint case, the semi-supervised learning result of data distillation (Table~\ref{tab:bdd_regime:small_coco}) is higher than that of fully-supervised training in \texttt{co-35}, but upper-bounded by that in \texttt{co-115}. However, in this case, the data distillation is closer to the lower bound (32.3 \vs 30.5) and farther away from the upper bound.
This result requires further exploration, which we leave to future work.

\paragraph{Large-scale data.} Table~\ref{tab:bdd_regime:large_coco} shows the data distillation result using \texttt{co-115} as labeled and \texttt{un-120} as unlabeled data, comparing with the fully-supervised result in \texttt{co-115}. Our method is able to improve over the fully-supervised baselines. Although the gains may appear small (0.8-0.9 points in AP and 0.9-1.1 points in AP$_{50}$), the signal is consistently observed for \emph{all network backbones and for all metrics}. The biggest improvement is seen in the AP$_M$ metric, with an increase of up to 1.8 points (from 43.7 to 45.5 in ResNeXt-101-32$\times$4).

The results in Table~\ref{tab:bdd_regime:small_coco} and \ref{tab:bdd_regime:large_coco} suggest that object detection with unlabeled data is a more challenging task, but unlabeled data with data distillation can still help.

\section{Conclusion}

We show that it is possible to surpass large-scale supervised learning with \emph{omni-supervised learning}, \ie, using all available supervised data together with large amounts of unlabeled data. We achieve this by applying \emph{data distillation} to the challenging problems of COCO object and keypoint detection. We hope our work will attract more attention to this practical, large-scale setting.

\section*{Acknowledgements}

We would like to thank Daniel Rueckert for his support and guidance during the initial stages of the project.

{
\small
\bibliographystyle{ieee}
\bibliography{distillation}
}

\end{document}